\documentclass{article}
\usepackage{fix-cm}

\PassOptionsToPackage{numbers, compress}{natbib}

\usepackage[preprint]{neurips_2025}

\usepackage[utf8]{inputenc} 
\usepackage[T1]{fontenc}    
\usepackage{url}            
\usepackage{amsfonts}       
\usepackage{nicefrac}       
\usepackage{microtype}      
\usepackage[table]{xcolor}

\definecolor{darkred}{RGB}{139,0,0}
\definecolor{darkblue}{RGB}{0,0,139}
\definecolor{lightblue}{RGB}{242,249,255}
\definecolor{lightpink}{RGB}{255,239,239}

\usepackage{floatrow}
\newfloatcommand{capbtabbox}{table}[][\FBwidth]
\usepackage{wrapfig}

\usepackage{caption}

\usepackage{amsmath}

\usepackage{hyperref}       
\usepackage{cleveref}
\crefname{figure}{Fig.}{Figs.}
\crefname{table}{Table.}{Tables.}
\crefname{section}{Section}{Secs.}

\usepackage{graphicx}
\usepackage{tabularx}
\usepackage{siunitx}
\usepackage{booktabs}       
\usepackage{multirow}
\usepackage{multicol}
\usepackage{makecell}

\usepackage{pifont}
\usepackage{amssymb}
\newcommand{\tabsize}{\fontsize{7.3pt}{10pt}\selectfont}
\newcommand{\titlesize}{\fontsize{8.5pt}{10pt}\selectfont}
\title{DualComp: End-to-End Learning of a Unified Dual-Modality Lossless Compressor}

\author{%
  Yan Zhao\textsuperscript{\rm 1}, Zhengxue Cheng\textsuperscript{\rm 1}, Junxuan Zhang\textsuperscript{\rm 2}, Qunshan Gu\textsuperscript{\rm 2}, Qi Wang\textsuperscript{\rm 2}, Li Song\textsuperscript{\rm 1} \\
  \textsuperscript{\rm 1}Institute of Image Communication and Network Engineering, Shanghai Jiao Tong University\\
  \textsuperscript{\rm 2}Ant Group, Hangzhou, China\\
  \textsuperscript{\rm 1}zhaoyanzy@sjtu.edu.cn, zxcheng@sjtu.edu.cn, song\_li@sjtu.edu.cn\\
  \textsuperscript{\rm 2}junxuan.zjx@antgroup.com, qunshan.gu@antgroup.com, qw.qq@antgroup.com
}

\begin{document}

\maketitle

\begin{abstract}

Most learning-based lossless compressors are designed for a single modality, requiring separate models for multi-modal data and lacking flexibility. However, different modalities vary significantly in format and statistical properties, making it ineffective to use compressors that lack modality-specific adaptations. 
While multi-modal large language models (MLLMs) offer a potential solution for modality-unified compression, their excessive complexity hinders practical deployment. To address these challenges, we focus on the two most common modalities, image and text, and propose \textbf{\emph{DualComp}}, \textbf{the first unified and lightweight learning-based dual-modality lossless compressor}. Built on a lightweight backbone, \emph{DualComp} incorporates three key structural enhancements to handle modality heterogeneity: modality-unified tokenization, modality-switching contextual learning, and modality-routing mixture-of-experts. A reparameterization training strategy is also used to boost compression performance. DualComp integrates both modality-specific and shared parameters for efficient parameter utilization, enabling near real-time inference ($200$KB/s) on desktop CPUs. With much fewer parameters, DualComp achieves compression performance on par with the SOTA LLM-based methods for both text and image datasets. Its simplified single-modality variant surpasses the previous best image compressor on the Kodak dataset by about $9\%$ using just 1.2\% of the model size. 
\end{abstract}

\section{Introduction}


Information theory \cite{codingtheory} proves that the minimum number of bits needed to represent data is determined by its $-\log{2}$ probability \cite{entropy}. This forms the basis for combining probabilistic modeling with entropy coding \cite{AC:1991, Huffman:1952, ANS:2013} to achieve lossless compression. Leveraging their contextual predictive strengths, neural networks, especially large language models (LLMs), have recently been combined with entropy coding to improve compression performance \cite{lmic,llmzip,tszip,p2llm,l3tc,msdzip,l3c,l3tc}.


Though notable progress has been made, some key limitations still remain. \underline{First}, these methods typically use models with billions of parameters, far exceeding the size of the data being compressed. As encoding and decoding speed roughly equals the prediction model's inference speed, this complexity becomes a major bottleneck. For instance, \citet{junhao} and \citet{p2llm} use fine-tuned LLaMA-8B models \cite{llama3} for lossless image compression, requiring over 30 minutes to decode a single 1080p image. \citet{nncpv2:2021} and \citet{cmix:2023} leverage Transformers \cite{transformer:2017} and LSTMs \cite{lstm:1997} for text compression, requiring 2.8 and 7.2 days, respectively, to decode just 1GB of text. \underline{Second}, most approaches support only a single modality, requiring separate compressors for image and text in multi-modal scenarios, which increases the system complexity and deployment costs. Though \citet{lmic} explores multi-modal compression, it neglects modality-specific characteristics by uniformly encoding all data as ASCII text, resulting in relatively poor performance on non-text modalities. In fact, different modalities exhibit distinct characteristics. Text naturally has a 1D sequential structure with intra- and inter-sentence semantic dependencies, while images are two-dimensional with strong spatial and inter-channel correlations. Ignoring these structural differences and simply feeding both modalities into an unadapted model would result in quite poor compression performance. 






\begin{figure}
    \centering
    \includegraphics[width=\linewidth]{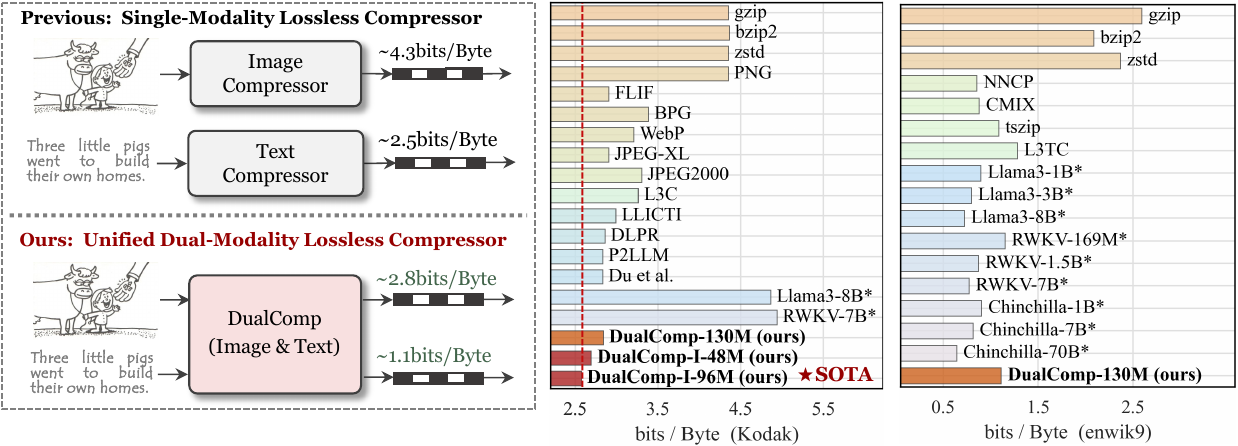}
    \vspace{-15pt}
    \caption{\titlesize Left: Most existing lossless compressors support only a single modality, whereas DualComp enables dual-modality compression in one model. Right: Lossless compression performance (bits/Byte) on image (Kodak) and text (enwik9) datasets. DualComp matches or surpasses SOTA methods with fewer parameters on both image and text.}
    \label{fig:teaser}
\end{figure}

To address these challenges, we propose DualComp, the first unified and lightweight dual-modality lossless compressor for both image and text. We select the latest RWKV-7 \cite{rwkv7} as our lightweight backbone and introduce three key enhancements for efficient dual-modality compression. \underline{First}, we adopt a shared vocabulary to achieve modality-unified tokenization for both image and text. \underline{Second}, considering the distinct contextual patterns of image and text, we separate their core contextual learning procedure with a modality-switching mechanism. \underline{Third}, we replace the model's final multilayer perceptron (MLP) \cite{mlp} with a modality-routing mixture-of-experts (MoE) architecture \cite{moe} to boost dual-modality representation. \underline{Besides}, inspired by \cite{l3tc}, we apply a reparameterization training strategy to improve compression performance without increasing inference complexity.


DualComp integrates both modality-specific and shared parameters, enabling efficient dual-modality lossless compression with fewer parameters than using separate single-modality models. It achieves a 57\% higher compression efficiency than gzip \cite{gzip} for text compression and outperforms PNG \cite{png} by 70\% for image compression. It matches the SOTA methods with quite fewer parameters and supports near real-time inference on desktop CPUs. Its single-modality variant surpasses previous best image compressors by 9\% with 1.2\% of the model size. Our contributions can be concluded as:

\begin{itemize}
    \item We propose DualComp, a unified and lightweight dual-modality lossless compressor for both image and text data. To our best knowledge, it is the first learning compressor to explicitly address modality heterogeneity and enable efficient dual-modality compression within a single framework. It provides useful insights for the emerging multi-modal applications.
    \item Built on a lightweight backbone, DualComp integrates three key enhancements for efficient dual-modality compression: modality-unified tokenization, modality-switching contextual learning, and modality-routing mixture-of-experts. A reparameterization training strategy is also utilized to boost compression performance without increasing complexity.

    \item DualComp matches or surpasses other SOTA compressors on both image and text datasets. It compresses the Kodak dataset up to 2.57 bits/Byte, outperforming the previous best method by about 9\% with only 1.2\% of the model size. Due to the parameter-efficient designs, it supports near real-time inference ($200$KB/s) on desktop CPUs.  

\end{itemize}



\section{Related Work}

\paragraph{Classical Lossless Compressors.}

\begin{figure}
    \centering
    \includegraphics[width=0.98\linewidth]{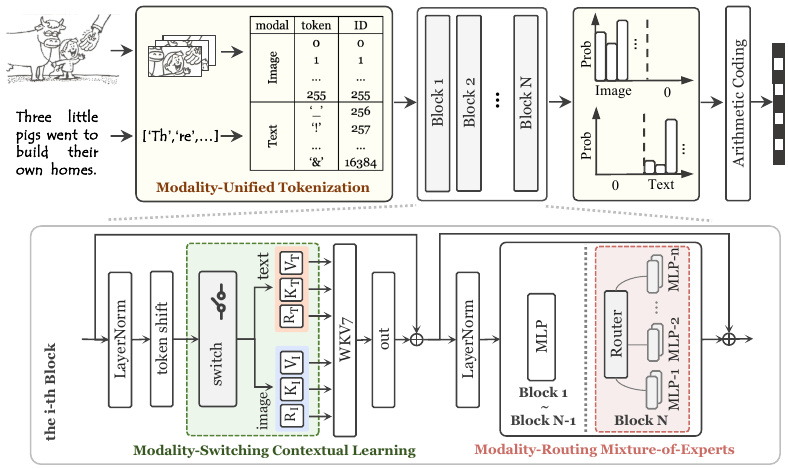}
    \vspace{-6pt}
    \caption{\titlesize DualComp tokenizes text and image inputs with a unified vocabulary, then encodes them into a compressed bitstream via context probabilities and arithmetic coding. Built on a lightweight backbone, it further incorporates modality-switching context learning and a modality-routing mixture-of-experts for efficient dual-modality compression.}
    \label{fig:overview}
\end{figure}
Classical lossless compression methods fall into two main categories: (1) general-purpose compressors such as gzip \cite{gzip}, bzip2 \cite{bzip2}, and zstd \cite{zstd}, and (2) modality-specific compressors tailored to particular data types. Gzip combines LZ77 \cite{lz77:1977} with Huffman coding \cite{Huffman:1952}, while bzip2 improves compression using the Burrows-Wheeler transform \cite{burrows-wheeler}, move-to-front encoding \cite{movetofront}, and Huffman coding. Zstd enhances efficiency by combining fast LZ77-style matching with finite-state entropy coding \cite{finitecoding}. General-purpose compressors are commonly used for text, while image compression relies on specialized methods that exploit spatial redundancy. Examples include PNG \cite{png}, WebP \cite{webp}, FLIF \cite{flif}, JPEG-XL \cite{jpegxl}, JPEG-2000 \cite{jpeg2000}, and BPG \cite{bpg}, which apply techniques like predictive coding, filtering, and transform-based compression. FLIF uses MANIAC entropy coding to adaptively model data distributions, PNG employs filtering and Deflate, and BPG leverages advanced intra-frame coding and CABAC \cite{cabac} for higher efficiency.


\paragraph{Learning-based Lossless Compressors.}



Learning-based compressors combine neural predictive modeling with entropy coding to outperform classical methods. However, due to the distribution sensitivity of neural models and distinct characteristics of different modalities, most learning-based compressors are designed for a single modality (image \cite{junhao, p2llm, l3c, dlpr,ivpf,iflow,rc} or text \cite{finezip, llmzip, nncpv2:2021, cmix:2023, l3tc, tszip}). For example, NNCP and CMIX use Transformer and LSTM models for text compression, while \citet{l3c} applies convolution networks for image compression. With the rise of LLMs, recent works have combined them with entropy coding for efficient lossless compression \cite{junhao, p2llm, finezip, llmzip,tszip,lmic}. While \citet{lmic} explores multi-modal compression, it ignores modality-specific structure by encoding all data as ASCII text, resulting in weak performance on non-text inputs. 

\section{Proposed Method}


As shown in \cref{fig:complexity}, the proposed DualComp enables lossless compression for the two most common modalities: text and image. Both are first tokenized into sequences $\{x_1, x_2, ..., x_n\}$ and processed by a probabilistic model to estimate contextual probabilities $p(x_i|x_{<i})$. Arithmetic coding (AC) \cite{AC:1991} is then used to compress the data near its entropy bound: $H(p) = \mathop{\mathbb{E}} (\sum_{i=1}^{n} -\log_{2} p(x_i|x_{<i})])$. 


\begin{figure}
    \centering
    \includegraphics[width=\linewidth]{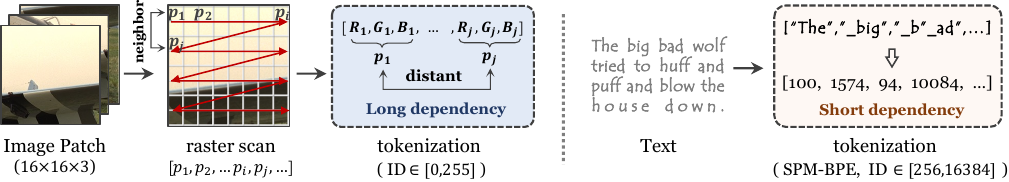}
    \vspace{-15pt}
    \caption{\titlesize Dual-modality tokenization: images are patched and scanned into 1D sequences, with each subpixel as a token. Text is tokenized using an SPM-BPE tokenizer. The two modalities share a unified vocabulary of 16K size.}
    \label{fig:tokenize}
\end{figure}

The compression complexity is mainly bounded by the model's inference speed and scales linearly with the number of tokens. To ensure efficient compression, we adopt the lightweight RWKV-7 model as the backbone of our DualComp framework (\cref{sec:choose}), and further introduce three structural modifications to better support dual-modality data compression: (1) We use a modality-unified tokenizer to generate text and image tokens (\cref{sec:tokenization}). (2) A modality-switching mechanism is integrated into the Time Mixing module to implement each modality's contextual learning (\cref{sec:gating}). (3) The model's final MLP is replaced with a modality-routing MoE architecture to boost dual-modality representation (\cref{sec:moe}). Besides, inspired by \cite{l3tc}, we apply a high-rank reparameterization strategy to boost representation capacity without increasing inference complexity (\cref{sec:hira}).

\subsection{Preliminaries: Low-Complexity Backbone}


\paragraph{Parameter-Efficient Attention Network.}
\label{sec:choose}

Using the method in \cite{lmic}, we compare three common language model architectures, Transformer \cite{transformer:2017}, Mamba \cite{mamba}, and RWKV-7 \cite{rwkv7}, on lossless image compression to identify a backbone that best balances compression performance and complexity.

\begin{wraptable}{r}{0.44\textwidth} 
\vspace{-8pt}
    \centering
    \tabsize
    \renewcommand{\arraystretch}{0.86}
    \setlength{\tabcolsep}{3.8pt}
    \caption{\titlesize Compression performance of Transformer and RWKV-7 on Kodak using the method in \cite{lmic}.}
    \begin{tabular}{lccc}
    \toprule
     \textbf{Model} & {\textbf{MACs}$\downarrow$} & {\textbf{bits/Byte}$\downarrow$} & \textbf{Speed (KB/s)} $\uparrow$ \\
    \midrule
    Tsfm-3.2M & 3.92M & 3.700 & $59.8$ \\
    Tsfm-12M  & 10.8M & 3.451 & $58.2$ \\
    \midrule
    RWKV-3.2M & 4.69M & 3.439 & $166$ \\
    RWKV-12M  & 11.5M & 3.313 & $90.8$ \\
    \bottomrule
    \end{tabular}
    \vspace{-5pt}
    \label{tab:archcomp}
\end{wraptable}
Following \cite{lmic}, images are converted to grayscale, split into 2048-byte chunks, and encoded as ASCII text. A SentencePiece BPE tokenizer \cite{bpe} is used to generate tokens for context-based probability prediction. All models are trained on ImageNet \cite{imagenet} and evaluated on the Kodak dataset \cite{kodak} using bits/byte as metrics (uncompressed baselines: 8 bits/byte). We also assess computational cost using MACs and average inference speed (KB/s) on a MacBook Air. As shown in \cref{tab:archcomp}, RWKV outperforms Transformer in both compression efficiency and inference speed, making it the preferred choice for our model’s backbone. Mamba is also evaluated, but excluded from further comparison due to its slow training and inference speed.

\paragraph{Reparameterization Training Strategy.}
\label{sec:hira}

Inspired by \cite{l3tc}, we incorporate a high-rank reparameterization strategy to enhance inference performance without increasing complexity. An additional branch is added to each R, K, and V projection layer in the Time Mixing module \cite{rwkv7}. To increase capacity, we apply high-rank matrix decomposition to the reparameterization branches. During inference, these branches are merged into the main path, preserving a compact single-path structure.

\subsection{Modality-Unified Tokenization}
\label{sec:tokenization}

Text and image differ significantly in data distributions: image sub-pixels are typically 8-bit values in the range [0, 255], while text exhibits much higher variability, requiring vocabularies of 16K–256K tokens. To enable dual-modality compression, we adopt a modality-unified tokenization strategy. 

As shown in \cref{fig:tokenize}, for text, we utilize a SentencePiece BPE \cite{bpe} tokenizer with a 16K vocabulary following \cite{l3tc}. For images, each image is first divided into $16\times16\times3$ patches. Within each patch, pixels are flattened in raster-scan order, with each pixel's RGB channels sequentially expanded as three consecutive sub-pixels (i.e., $R_1,G_1,B_1,R_2,G_2,B_2,...$). This preserves both inter-channel and inter-pixel correlations, with each sub-pixel treated as an individual token, yielding a vocabulary size of 256. Subsequently, text and image vocabularies are merged into a unified token set for probability prediction, as depicted in \cref{fig:overview}. However, before arithmetic coding, we apply modality-specific masking to zero out non-target modality probabilities, thereby reducing estimation errors and improving compression efficiency. For example, when compressing images, logits associated with text tokens are zeroed out before applying softmax, and vice versa for text compression.


\begin{figure}[tb]
    \centering
    \includegraphics[width=\linewidth]{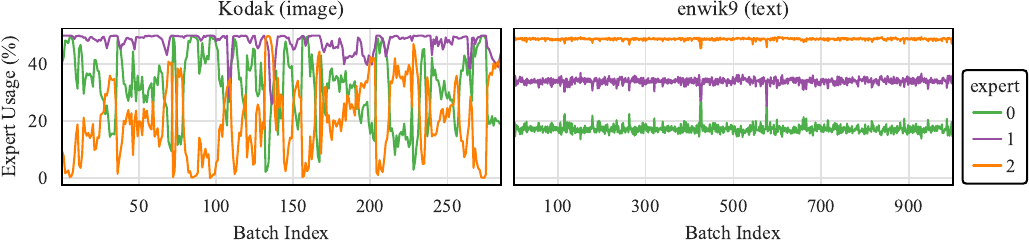}
    \vspace{-15pt}
    \caption{\titlesize Percent of expert usage in each batch when compressing image (left) and text (right) using DualComp-16M.}
    \label{fig:expert-usage}
\end{figure}

\subsection{Modality-Switching Contextual Learning}
\label{sec:gating}


Text and image modalities exhibit distinct contextual dependencies. Text follows a natural 1D sequential structure with intra- and inter-sequence semantic dependencies. In contrast, images are inherently two-dimensional and characterized by inter-pixel and inter-channel correlations. When flattened into 1D sequences via patching and raster scan, spatially adjacent pixels may become distant, resulting in more complex and longer-range dependencies than those in text, as shown in \cref{fig:tokenize}. These differences lead to distinct contextual probability patterns for images and text. Since the compressed bit rate closely approximates the entropy of predicted probabilities, precise contextual modeling is crucial. Hence, DualComp separates the core contextual learning parameters into modality-specific branches and uses a modality-switching mechanism to control their activations.




Specifically, in our RWKV backbone, each block consists of two components: a Time Mixing module and an MLP layer. The former primarily models contextual dependencies using operations like token shift, receptance/key/value (R/K/V) projection, WKV calculation, and state update, while the latter mainly enhances the nonlinear representation. To support modality-specific contextual patterns, we integrate a modality-switching mechanism into the Time Mixing module and maintain separate sets of R, K, V layers for text and images. The model dynamically switches between them based on the input modality, as illustrated in \cref{fig:overview}. As the R, K, V layers are simple linear projections, the introduced modality-specific parameters add little overhead but bring notable gains in dual-modality performance. The output layer of the Time Mixing module and the subsequent MLP layer do not encode temporal information. Thus, they are shared across modalities to maximize parameter utilization efficiency.

\subsection{Modality-Routing Mixture-of-Experts}
\label{sec:moe}

MoE is a sparse architecture that routes each token to selected experts, enabling greater model capacity and flexibility. To further improve dual-modality representation capability, we replace the MLP layer in the final block with an MoE layer to perform learnable modality-routing. 

Specifically, as shown in \cref{fig:overview}, the original large MLP in the final block is replaced by multiple smaller expert MLPs. A learnable router assigns each token $x_i$ a routing score $g_{i,e}$ for each expert $e$, and activates the top-$k$ experts with the highest scores to deal with the token in parallel. Their outputs are aggregated via a weighted sum as $\text{MoE}(x_i) = \sum_{e\in \text{top}-k}\hat{g}_{i,e}\cdot e(x_i)$, where $\hat{g}_{i,e}$ is the re-normalized score. To maintain a lightweight design, DualComp uses a compact MoE with three experts, each about half the size of the original large MLP, and activates the top two per token. Despite its small size, the modality-routing MoE notably enhances dual-modality compression by introducing additional learnable modality-specific behaviors that help the model adapt to the distinct properties of image and text. Using the 16M DualComp model as an example, \cref{fig:expert-usage} shows expert usage per batch (\%) for image and text compression. Image batches show higher variability, prompting more dynamic expert routing, while text batches are more uniform, resulting in more consistent expert selection.




\begin{table*}[tb]
    \belowrulesep=0pt
    \aboverulesep=0pt
    \centering
    \tabsize
    \renewcommand{\arraystretch}{0.96}
    \setlength{\tabcolsep}{2.35pt}
    \newcommand{\centerdash}[1]{\ifx#1-\multicolumn{1}{c}{-}\else#1\fi}
    \vspace{-6pt}
    \begin{tabular}
    {p{0.28cm}p{1.7cm}|p{1.22cm}cc|c|p{1.42cm}p{1.42cm}p{1.42cm}p{1.42cm}|c}
    \toprule
     & \multirow{2}{*}{\textbf{Compressor}} 
     & \multirow{2}{*}{\textbf{\#Params}$\downarrow$} 
     & \multirow{2}{*}{\textbf{MACs}$\downarrow$} 
     & \multirow{2}{*}{\makecell{\textbf{Speed}$\uparrow$\\ \textbf{(KB/s)}}}
     & \multirow{2}{*}{\textbf{\makecell{Dual\\Modal}}} 
     & \multicolumn{4}{c}{\textbf{bits/Byte$\downarrow$\quad [image]}}
     & \textbf{[text]} \\[1pt]
     \cmidrule{7-11}

     & & & & & 
     & \textbf{Kodak} & \textbf{CLIC-P} & \textbf{CLIC-M} & \textbf{DIV2K} & \textbf{enwik9} \\
    \midrule
    \multirow{9}{*}{\rotatebox{90}{\textbf{Classical}}} 
    & gzip$^\dag$ \cite{gzip}   & - & - & - & \checkmark 
    & 4.349 & 4.039 & 3.947 & 4.224 & 2.590\\
    & bzip2$^\dag$ \cite{bzip2} & - & - & - & \checkmark
    & 4.359 & 4.033 & 3.931 & 4.208 & 2.082 \\
    & zstd$^\dag$ \cite{zstd}   & - & - & - & \checkmark
    & 4.350 & 4.042 & 3.952 & 4.229 & 2.364 \\
    & FLIF$^\dag$ \cite{flif}   & - & - & - & \ding{55} 
    & 2.903 & 2.792 & 2.497 & 2.914 &  - \\
    & BPG$^\dag$ \cite{bpg}     & - & - & - & \ding{55} 
    & 3.382 & 3.144 & 2.855 & 3.285 &  - \\
    & WebP$^\dag$ \cite{webp}   & - & - & - & \ding{55} 
    & 3.205 & 3.006 & 2.774 & 3.176 &  - \\
    & JPEG-XL$^\dag$ \cite{jpegxl}  & - & - & - & \ding{55} 
    & 2.902 & 2.722 & 2.395 & 2.826 &  - \\
    & JPEG2000$^\dag$ \cite{jpeg2000} & - & - & - & \ding{55} 
    & 3.301 & 2.927 & 2.667 & 3.127 &  - \\
    & PNG$^\dag$ \cite{png}     & - & - & - & \ding{55} 
    & 4.349 (+54\%) & 4.041 (+72\%) & 3.952 (+90\%) & 4.229  (+68\%) &  - \\
    \cmidrule{1-11}

    \multirow{12}{*}{\rotatebox{90}{\textbf{Learning-based}}}
    & NNCP \cite{nncpv2:2021} & - & 187M & 1.6 & \ding{55} 
    & - & - & - & - & \underline{0.853} \\
    & CMIX \cite{cmix:2023} & - & - & 4 & \ding{55} 
    & - & - & - & - & \underline{0.879} \\
    & tszip \cite{tszip}  & 169M & 131M & 180$^\ddag$ & \ding{55} 
    & - & - & - & - &  1.083 \\
    & L3TC \cite{l3tc}  & 12M & 13.0M & 580$^\ddag$ & \ding{55} 
    & - & - & - & - &  1.280 \\
    
    & L3C \cite{l3c}     & 5M & 6.86M & 178 & \ding{55} 
    & 3.260  & 2.940 & 2.640 & 3.090 & - \\
    & RC \cite{rc}       & - & - & - & \ding{55} 
    & - & 2.930 & 2.540 & 3.080 & - \\
    & iVPF \cite{ivpf}   & - & - & - & \ding{55} 
    & - & 2.540 & 2.390 & 2.680 & - \\
    & iFlow \cite{iflow} & - & - & - & \ding{55} 
    & - & \underline{2.440} (+4\%) & 2.260 & 2.570 & - \\
    & DLPR \cite{dlpr}   & - & - & 640  & \ding{55} 
    & 2.860  & \underline{2.380} (+1\%) & \underline{2.160} (+4\%) & \underline{2.550} (+2\%) & - \\
    & P2LLM \cite{p2llm} & 8B ($1\times$) & - & 3$^\ddag$ & \ding{55} 
    & \underline{2.830} (0\%) 
    & \textbf{2.350} (0\%) 
    & \textbf{\color{darkblue} 2.080} (0\%) 
    & \textbf{\color{darkblue} 2.510} (0\%) 
    & - \\

    & $\text{Llama3}^{*\dag}$ \cite{lmic} & 8B & 7.80G & 3$^\ddag$ & \checkmark & 4.862 & 4.290 & 4.234 & 4.378 & \textbf{\color{darkblue}0.722} \\

    & $\text{RWKV}^{*\dag}$ \cite{lmic}  & 7B & 7.19G  & 5$^\ddag$ & \checkmark & 4.937 & 4.310 & 4.298 & 4.718 & \textbf{0.774} \\
    \cmidrule{1-11}

    \multirow{7}{*}{\rotatebox{90}{\textbf{Proposed}}}
    & \cellcolor{lightblue}DualComp-I  
    & \cellcolor{lightblue}12M 
    & \cellcolor{lightblue}11.6M 
    & \cellcolor{lightblue}614$^\ddag$
    & \cellcolor{lightblue}\ding{55}
    & \cellcolor{lightblue}\underline{2.823} (-0\%) 
    & \cellcolor{lightblue}2.674 
    & \cellcolor{lightblue}2.448  
    & \cellcolor{lightblue}2.862 
    & \cellcolor{lightblue}- \\
    
    & \cellcolor{lightblue}DualComp-I  
    & \cellcolor{lightblue}48M ({\color{darkblue}$\frac{1}{167}$}) 
    & \cellcolor{lightblue}33.3M 
    & \cellcolor{lightblue}360$^\ddag$
    & \cellcolor{lightblue}\ding{55} 
    & \cellcolor{lightblue}\textbf{2.693} ({\color{darkblue}-5\%}) 
    & \cellcolor{lightblue}2.455 
    & \cellcolor{lightblue}\underline{2.193} (+5\%) 
    & \cellcolor{lightblue}\underline{2.627} (+5\%)
    & \cellcolor{lightblue}- \\
    
    & \cellcolor{lightblue}DualComp-I   
    & \cellcolor{lightblue}96M ({\color{darkblue}$\frac{1}{83}$}) 
    & \cellcolor{lightblue}59.9M
    & \cellcolor{lightblue}317$^\ddag$
    & \cellcolor{lightblue}\ding{55}
    & \cellcolor{lightblue}\textbf{\color{darkblue} 2.571} ({\color{darkblue}-9\%}) 
    & \cellcolor{lightblue}\textbf{\color{darkblue} 2.350} (-0\%) 
    & \cellcolor{lightblue}\textbf{2.110} (+1\%) 
    & \cellcolor{lightblue}\textbf{2.547} (+2\%) 
    & \cellcolor{lightblue}- \\
    \cmidrule{2-11}

    & \cellcolor{lightpink}DualComp   
    & \cellcolor{lightpink}16M 
    & \cellcolor{lightpink}12.6M 
    & \cellcolor{lightpink}550$^\ddag$
    & \cellcolor{lightpink}\color{darkblue}{\checkmark}
    & \cellcolor{lightpink}3.026 
    & \cellcolor{lightpink}2.691 
    & \cellcolor{lightpink}2.498
    & \cellcolor{lightpink}2.901 
    & \cellcolor{lightpink}1.218 \\
    
    & \cellcolor{lightpink}DualComp  
    & \cellcolor{lightpink}66M ({\color{darkblue}$\frac{1}{121}$}) 
    & \cellcolor{lightpink}36.0M 
    & \cellcolor{lightpink}280$^\ddag$
    & \cellcolor{lightpink}\color{darkblue}{\checkmark} 
    & \cellcolor{lightpink}2.889 
    & \cellcolor{lightpink}2.533 
    & \cellcolor{lightpink}2.370
    & \cellcolor{lightpink}2.741 
    & \cellcolor{lightpink}1.135 \\
    
    & \cellcolor{lightpink}DualComp  
    & \cellcolor{lightpink}130M ({\color{darkblue}$\frac{1}{60}$}) 
    & \cellcolor{lightpink}64.0M 
    & \cellcolor{lightpink}195$^\ddag$
    & \cellcolor{lightpink}\color{darkblue}{\checkmark}
    & \cellcolor{lightpink}\underline{2.834} (+0\%) 
    & \cellcolor{lightpink}\underline{2.364} (+0\%)  
    & \cellcolor{lightpink}\underline{2.155} (+3\%)
    & \cellcolor{lightpink}\underline{2.554} (+2\%)
    & \cellcolor{lightpink}1.107 \\
    
    \bottomrule
    \end{tabular}
    \caption{\titlesize Comprehensive lossless compression results on image and text datasets: best (bits/Byte) in \textbf{\color{darkblue}bold blue}, second in \textbf{bold}, and third to fifth \underline{underlined}. Image compression gains and model size ratios over the previous best method \cite{p2llm} are reported as percents (\%). $*$ denote pretrained LLMs. $\dag$ marks the methods we reprodeuced, while $\ddag$ presents the inference speeds measured on GPU@A100. All other unmarked values are claimed by their original papers. 
}
    \label{tab:compare}
\end{table*}





\subsection{Loss Function}

Our loss function comprises two terms: a cross-entropy loss and an auxiliary MoE loss, as shown in \cref{equ:loss}. The cross-entropy loss measures the divergence between the predicted and target distributions. The auxiliary loss promotes balanced expert utilization by penalizing variance in expert importance and token distribution, where $\mathrm{CV}^2$ denotes the squared coefficient of variation, $g_{i,e}$ is the routing score of the $i$-th token to the $e$-th expert. The two losses are weighted with $\lambda=0.01$.

\begin{equation}
\mathcal{L} = \underbrace{-\sum q \log p}_{\text{Cross Entropy}} + \lambda \Big[ 
\underbrace{\mathrm{CV}^2(\sum g_{i,e})}_{\text{Expert Importance}} + 
\underbrace{\mathrm{CV}^2(\sum \mathbb{I}(g_{i,e}>0))}_{\text{Load Balance}} \Big]
\label{equ:loss}
\end{equation}



\section{Experiments}

\subsection{Experimental Setup}

\paragraph{Implementation Details.}


To vary the model size, we adjust the number of layers and the embedding dimension. Our model parameters are divided into two groups: modality-specific and modality-shared. To optimize both parameter groups effectively, we use the FusedAdam optimizer \cite{fusedadam} and adopt a \underline{three-stage training strategy}: (1) freeze the shared parameters and train only the modality-specific ones for two epochs with a fixed learning rate of $2\times 10^{-5}$. (2) freeze the modality-specific parameters and train the shared ones for two epochs at a learning rate of $2\times 10^{-5}$. (3) unfreeze all parameters and train the full model for 16 epochs with a cosine annealing learning rate scheduler \cite{scheduler}, starting at $1\times 10^{-4}$ and decaying to $5\times 10^{-6}$. The training is performed on two NVIDIA A100 GPUs. 


We also design a simplified single-modality variant, \underline{DualComp-I}, by removing the dual-modality proposals from DualComp and retaining only the image tokenization and reparameterization training. DualComp-I is trained following the same procedure as the third stage of DualComp. 


\paragraph{Dual-Modality Datasets.} 

Our dataset includes both image and text modalities. The training set consists of 5,500 images from ImageNet \cite{imagenet} and 100MB of text from enwik8 \cite{enwik8}. 
During training, we mix image and text data within each batch at a 1:1 ratio to ensure balanced and stable learning. For evaluation, we use Kodak \cite{kodak}, CLIC-Mobile, CLIC-Pro \cite{clic}, and DIV2K \cite{div2k} as image test sets, and enwik9 \cite{enwik9} (1GB of text) as the text test set. 


\begin{figure}
    \centering
    \includegraphics[width=\linewidth]{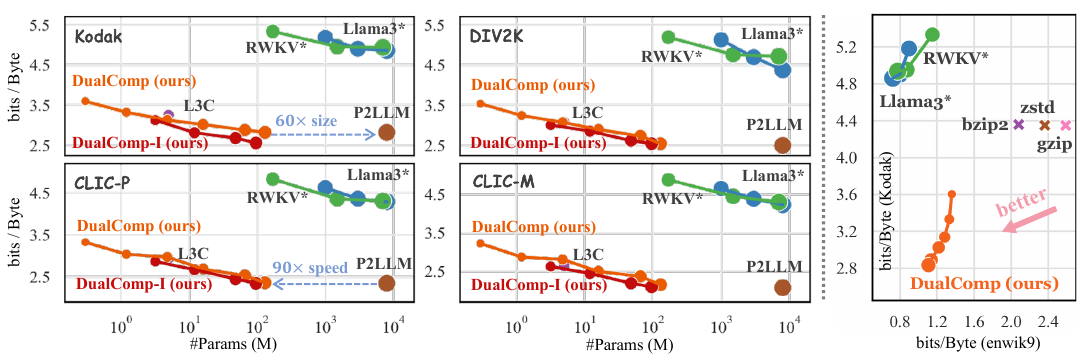}
    \vspace{-15pt}
    \caption{\titlesize Left: Learning-based methods' image compression performance (bits/Byte vs. model size). The closer to the bottom-left corner, the smaller the model and the better the performance. Right: Dual-modality compression consistency. The x-axis and y-axis are bits/Byte on text and image datasets, respectively. $*$ marks pretrained LLMs.}
    \label{fig:curves}
\end{figure}

\paragraph{Compared Methods and Metrics.}

We compare lossless compression methods for both text and image modalities, including classical and learning-based approaches. Classical baselines include general-purpose compressors (gzip \cite{gzip}, bzip2 \cite{bzip2}, zstd \cite{zstd}) and image-specific methods (PNG \cite{png}, WebP \cite{webp}, FLIF \cite{flif}, JPEG XL \cite{jpegxl}, JPEG2000 \cite{jpeg2000}, BPG \cite{bpg}). For learning-based methods, we evaluate recent image lossless compressors \cite{dlpr, p2llm, l3c, rc, ivpf, iflow, llicti} and text lossless compressors \cite{nncpv2:2021, cmix:2023,l3tc,tszip}. We also compare with the only learning-based dual-modality compressor \cite{lmic} that uses pre-trained LLMs like Chinchilla \cite{chinchilla}, LLaMA3 \cite{llama3}, and RWKV \cite{rwkv7}.


%

We evaluate compression performance for both image and text using bits/Byte as metrics. Model complexity is assessed via MACs and inference speed (KB/s), measured on both a high-performance NVIDIA A100 GPU server and a lightweight desktop device (MacBook Air).


\subsection{Dual-Modality Lossless Compression Performance}


\paragraph{Image Compression Performance.}

\Cref{fig:teaser}, \cref{tab:compare}, and \cref{fig:curves} compare our methods against existing lossless image compression approaches. Herein, we not only evaluate the proposed DualComp but also its simplified single-modality version, DualComp-I.

Among classical compressors, general-purpose methods perform poorly on images, with bits/byte values around 4. Image-specific compressors like FLIF and JPEG-XL perform obviously better, offering about 30\% improvements over gzip. Learning-based approaches leverage neural networks for greater efficiency. Except for \cite{lmic}, existing methods \cite{l3c,rc,ivpf,iflow,dlpr,p2llm,junhao} all serve as image-specific compressors. P2LLM \cite{p2llm} achieves efficient image compression using fine-tuned LLaMA3-8B models but incurs excessive training and inference complexity. It requires about half an hour to decode a single 1080p image. Although \citet{lmic} supports dual-modality compression, it neglects the differences between modalities by simply converting images to ASCII text and using pretrained LLMs for probability prediction. Consequently, despite having billions of parameters, it derives poor image compression performance (even 10\% worse than gzip).

By contrast, the proposed DualComp reaches comparable compression efficiency to SOTA approaches with much fewer parameters and faster inference, as shown in \cref{fig:curves}. Particularly, DualComp-I reaches SOTA results on the Kodak dataset and outperforms the previous best method by approximately $9\%$ using about 1.2\% of the parameters. While DualComp performs slightly below DualComp-I due to its shared capacity for both text and image, it still ranks top-three on all image datasets. Even with a lightweight 16M model, DualComp still surpasses PNG by nearly 50\%. Its stronger performance on Kodak is likely due to the training data being primarily low-resolution images.


\paragraph{Text Compression Performance.}

As shown in \cref{fig:teaser} and \cref{tab:compare}.
Classical general-purpose compressors like gzip, zstd, and bzip2 are typically used for text compression, achieving moderate bits/Byte between 2.082 and 2.590 on the enwik9 dataset. NNCP \cite{nncpv2:2021} and CMIX \cite{cmix:2023} were previously the best lossless text compressors \cite{benchmark} with compressed bits/Byte less than 0.88. However, they require 2.8 and 7.2 days, respectively, to decode the enwik9 dataset. Using the method in \cite{lmic}, pretrained LLMs such as LLaMA3-8B \cite{llama3} and RWKV-7B \cite{rwkv7} achieve superior text compression performance (less than 0.8 bits/Byte) with significantly high complexity. They require about 10 and 4 days, respectively, to decode the enwik9 dataset on an NVIDIA A100 GPU.



By contrast, our DualComp uses a lightweight model to perform dual-modality lossless compression. Though with two-thirds of parameters shared across modalities, it still outperforms L3TC \cite{l3tc}, a recent practical text compressor at similar model sizes. When using a 130M model, DualComp compresses enwik9 to 1.107 bits/Byte, yielding about 57\% gains over gzip. Besides, DualComp decodes the enwik9 dataset in less than two hours and is nearly $90\times$ faster than Llama3-8B.


\begin{table}[tb]
    \centering
    \tabsize
    \belowrulesep=0pt
    \aboverulesep=0pt
    \renewcommand{\arraystretch}{0.98}
    \setlength{\tabcolsep}{5.4pt}
    \vspace{-6pt}
    \begin{tabular}{c|cccc|ccc|cc}
    \toprule
      \textbf{Model} & \textbf{Reparam} & \textbf{Switching} & \textbf{MoE} & \textbf{Dual-Modal} 
      & \textbf{\#Params}$\downarrow$ & \textbf{MACs}$\downarrow$ & \textbf{Speed (KB/s)} $\uparrow$
      & \textbf{Kodak}$\downarrow$ & \textbf{enwik9}$\downarrow$ \\
      \midrule
      \multirow{6}{*}{\rotatebox{90}{\textbf{small}}} 
      & \ding{55}   & \ding{55}   & \ding{55}  & \ding{55}  & 219K & 0.85M & 151 & 3.739  & -  \\
      & \cellcolor{lightblue}\checkmark  & \cellcolor{lightblue}\ding{55}   
      & \cellcolor{lightblue}\ding{55}  & \cellcolor{lightblue}\ding{55}  
      & \cellcolor{lightblue}219K & \cellcolor{lightblue}0.85M & \cellcolor{lightblue}151
      & \cellcolor{lightblue}3.540  &  \cellcolor{lightblue}- \\
      \cmidrule{2-10}
      
      & \checkmark  & \ding{55}   & \ding{55}  & \checkmark & 219K & 0.85M & 151 & 3.970  &  1.987 \\
      & \checkmark  & \checkmark  & \ding{55}  & \checkmark & 273K & 0.85M & 140 & 3.587  &  1.794 \\
      & \checkmark  & \ding{55}   & \checkmark & \checkmark & 257K & 0.87M & 136 & 3.821  &  1.876 \\
      & \cellcolor{lightpink}\checkmark  & \cellcolor{lightpink}\checkmark  
      & \cellcolor{lightpink}\checkmark & \cellcolor{lightpink}\checkmark 
      & \cellcolor{lightpink}312K & \cellcolor{lightpink}0.87M & \cellcolor{lightpink}134
      & \cellcolor{lightpink}3.604  &  \cellcolor{lightpink}1.360 \\
      \midrule
      
      \multirow{6}{*}{\rotatebox{90}{\textbf{medium}}} 
      & \ding{55}   & \ding{55}   & \ding{55}  & \ding{55}  & 12M & 11.57M & 36 & 3.148 &  - \\
      & \cellcolor{lightblue}\checkmark & \cellcolor{lightblue}\ding{55}   
      & \cellcolor{lightblue}\ding{55}  & \cellcolor{lightblue}\ding{55}  
      & \cellcolor{lightblue}12M    & \cellcolor{lightblue}11.57M & \cellcolor{lightblue}36
      & \cellcolor{lightblue}2.822  &  \cellcolor{lightblue}- \\
      \cmidrule{2-10}
      & \checkmark  & \ding{55}   & \ding{55}  & \checkmark & 12M & 11.57M & 21 & 3.261  &  1.412 \\
      & \checkmark  & \checkmark  & \ding{55}  & \checkmark & 15M & 11.57M & 31 & 3.006  &  1.286 \\
      & \checkmark  & \ding{55}   & \checkmark & \checkmark & 14M & 12.55M & 24 & 3.335  &  1.300 \\
      & \cellcolor{lightpink}\checkmark  & \cellcolor{lightpink}\checkmark  
      & \cellcolor{lightpink}\checkmark & \cellcolor{lightpink}\checkmark 
      & \cellcolor{lightpink}16M & \cellcolor{lightpink}12.55M & \cellcolor{lightpink}22
      & \cellcolor{lightpink}3.026  &  \cellcolor{lightpink}1.218 \\
      \bottomrule
    \end{tabular}
    \caption{\titlesize Discussion on modality-switching contextual learning, modality-routing MoE, and reparameterization training. Compressed bits/Byte on Kodak and enwik9 is evaluated. Blue and pink rows indicate the proposed DualComp-I and DualComp models, respectively. Inference speeds (KB/s) are measured on a desktop CPU with a batch size of 128.}
    \label{tab:arch-ablations}
\end{table}

\begin{figure}
    \centering
    \includegraphics[width=0.98\linewidth]{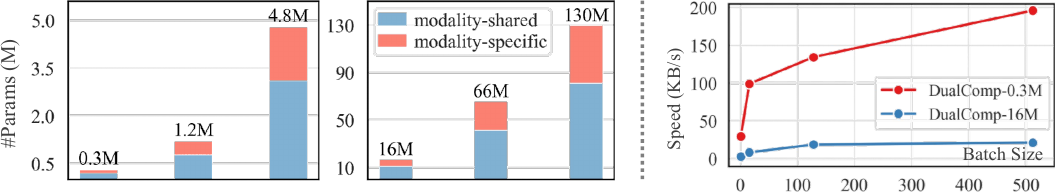}
    \vspace{-6pt}
    \caption{\titlesize Left: Parameter breakdown of DualComp at different scales (modality-specific vs. modality shared). Right: DualComp's inference speed (KB/s) on a MacBook Air CPU at varying batch sizes (1, 16, 128, and 512).}
    \label{fig:complexity}
\end{figure}



\paragraph{Dual-Modality Compression Consistency.}

\cref{fig:curves} depicts the dual-modality compression consistency on image (Kodak) and text (enwik9) datasets. Pretrained Llama3 and RWKV models follow the approach in \cite{lmic} and simply convert images to ASCII text. Due to the neglect of different modalities' characteristics, they perform quite poorly on image compression with compressed bits/Byte greater than 4.8. Classical compressors such as gzip, bzip2, and zstd are modality-agnostic but offer only moderate performance across both modalities. In contrast, DualComp introduces modality-specific designs and achieves effective compression on both image and text, reaching 2.834 bits/Byte on Kodak and 1.107 bits/Byte on enwik9 with compact models.

\subsection{Parameter Breakdown and Complexity Analysis}

DualComp includes both modality-specific and shared parameters, with the former handling modality heterogeneity and the latter enhancing representation for both modalities. As illustrated in \cref{fig:complexity}, across all proposed model scales (from 0.3M to 130M), the modality-specific parameters make up only one-third of the total, while the majority are shared. This design enables much more efficient parameter utilization compared to using two separate single-modality compressors. 

As for inference complexity, we evaluate DualComp on both desktop CPUs and server GPUs. As depicted on the right of \cref{fig:complexity}, on a MacBook Air CPU, the 0.3M (small) and 16M (medium) models reach up to 200KB/s and 25KB/s with a batch size of 512, respectively, enabling near real-time performance. \cref{tab:compare} further compares the learning-based methods' inference speeds on GPU servers. Our medium and large models reach fast inference speeds ranging from 195KB/s to 614KB/s on a GPU@A100 server with a batch size of 128, which are similar to other practical compressors like \cite{l3c, l3tc, tszip,dlpr} and approximately $90\sim200\times$ faster than those LLM-based methods \cite{lmic,p2llm}.

\subsection{Ablation Studies}
\label{sec：abaltion}

\paragraph{Discussion on Dual-Modality Proposals.}

We conduct ablation studies using the 0.3M (small) and 16M (medium) models to evaluate the effectiveness of our proposals for dual-modality processing, including the modality-unified tokenization, modality-switching contextual learning, and modality-routing MoE. As shown in \cref{tab:arch-ablations}, compared to the baseline without modality-specific design, introducing the modality-switching mechanism and the modality-routing MoE improves the dual-modality compression performance (bits/Byte on Kodak and enwik9 datasets) by about 10\% and 4\%, respectively, with model size increasing by no more than 25\% and negligible impact on inference speed (KB/s measured on a desktop CPU). Combining both strategies further improves overall compression and achieves a better balance between image and text modalities. Additionally, with modality-unified tokenization, masking non-target logits before probability output obviously improves compression performance on image compression and brings up to 20\% gains. 


\begin{table}[!tb]
    \centering
    \tabsize
    \belowrulesep=0pt
    \aboverulesep=0pt
    \renewcommand{\arraystretch}{0.98}
    \setlength{\tabcolsep}{8.8pt}
    \vspace{-6pt}
    \begin{tabular}{cccc|cc|c|cc}
    \toprule
    \multirow{2}{*}{\textbf{MoE layers}} 
    & \multirow{2}{*}{\textbf{Top-K}} 
    & \multirow{2}{*}{\textbf{Experts}} 
    & \multirow{2}{*}{\textbf{hidden factor}} 
    & \multicolumn{2}{c|}{\textbf{\#Params}$\downarrow$} 
    & \multirow{2}{*}{\textbf{MACs}$\downarrow$} 
    & \multicolumn{2}{c}{\textbf{bits/Byte$\downarrow$}}\\[1pt]

    \cmidrule{5-6}\cmidrule{8-9}
    & & & & \textbf{Total} & \textbf{Activated} & & \textbf{Kodak} & \textbf{enwik9} \\
    \midrule
    
    1 & 1 & 3 & $2\times$ & 312K & 294K & 0.87M & 3.657 & 1.458 \\
    \rowcolor{lightpink}
    1 & 2 & 3 & $2\times$ & 312K & 294K & 0.87M & 3.604 & 1.360 \\
    2 & 2 & 3 & $2\times$ & 349K & 313K & 0.89M & 3.616 & 1.333 \\
    2 & 2 & 4 & $2\times$ & 385K & 313K & 0.93M & 3.633 & 1.357 \\
    \cmidrule{1-9}
    1 & 2 & 3 & $0.5\times$ & 231K & 222K & 0.83M & 3.751 & 1.589 \\
    1 & 2 & 3 & $1\times$   & 258K & 240K & 0.85M & 3.681 & 1.432 \\
    \rowcolor{lightpink}
    1 & 2 & 3 & $2\times$   & 312K & 294K & 0.87M & 3.604 & 1.360 \\
    1 & 2 & 3 & $4\times$   & 420K & 348K & 0.93M & 3.591 & 1.341 \\
    
    \bottomrule
    \end{tabular}
    \caption{\titlesize Modality-routing MoE configuration analysis using the DualComp-0.3M model. The top part examines the expert selection strategy and number of MoE layers, while the bottom part evaluates the effect of expert size.}
    \label{tab:moe-ablation}
\end{table}

\begin{figure}
    \centering
    \vspace{-6pt}
    \includegraphics[width=\linewidth]{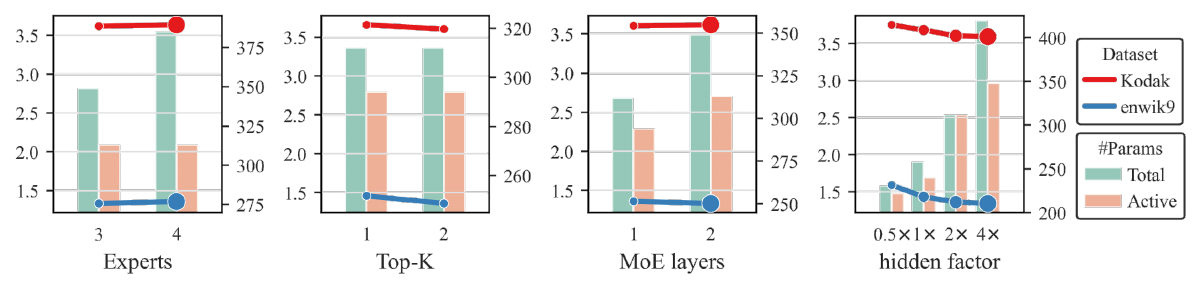}
    \vspace{-18pt}
    \caption{\titlesize Discussion on modality-routing MoE configurations using the DualComp-0.3M model. The line plots (left y-axis) indicate compression performance (bits/Byte) on image (Kodak) and text (enwik9) datasets, while the bar charts (right y-axis) present the total and activated parameters when processing a token. Detailed statistics are in \cref{tab:moe-ablation}.}
    \label{fig:ablation}
\end{figure}

\paragraph{Discussion on Reparameterization and MoE Configurations.}

We evaluate the impact of reparameterization training using the small and medium DualComp-I models, as shown in \cref{tab:arch-ablations}. It brings about 10\% improvements in compressed bits/Byte on the Kodak dataset without increasing model size or inference complexity. Additionally, we investigate the effect of different MoE configurations using the 0.3M DualComp model, as shown in \cref{tab:moe-ablation} and \cref{fig:ablation}. It can be seen that replacing only the final MLP with an MoE achieves similar performance to replacing all MLPs (equaling 2 in this model), but introduces obviously lower complexity. Likewise, increasing the number of experts from three to four or reducing the Top-K values shows no clear benefit in compression performance. Therefore, we adopt the most parameter-efficient configuration, as highlighted in \cref{tab:moe-ablation}. We also evaluate the effect of expert sizes by adjusting the hidden dimension of each MLP expert (bottom part of \cref{tab:moe-ablation}). The performance saturates at a hidden factor of 2, with minimal gains beyond that. Since the original MLP uses a hidden factor of 4, this MoE design introduces little parameter overhead.






\section{Conclusion}

We propose DualComp, the first unified and lightweight lossless compressor for image and text. Built on a lightweight backbone, DualComp integrates three key enhancements to deal with modality heterogeneity: modality-unified tokenization, modality-switching contextual learning, and modality-routing mixture-of-experts. A reparameterization training strategy is also adopted to boost performance without increasing inference complexity. On both image and text compression, DualComp achieves compression performance comparable to the SOTA methods with much fewer parameters and supports near real-time inference on desktop CPUs. Its simplified single-modality variant, DualComp-I, outperforms the previous best image compressors by about 9\% with about 1.2\% parameters. In future works, we will explore more modalities and develop a more unified multi-modal lossless compressor.

{
\footnotesize
\bibliographystyle{unsrtnat}
\bibliography{ref}
}

\clearpage

\section*{Supplementary Material}

The supplementary material provides additional implementation details and experimental results omitted from the main paper due to space constraints. It offers both deeper explanations of the proposed method and extended experiments, including: (1) detailed model configurations, (2) introductions of the DualComp-I variant, (3) explanations of the reparameterization training strategy, (4) descriptions of the modality-routing mixture-of-experts. (5) details of the dual-modality datasets, (6) extended lossless compression performance analyses, (7) additional inference speed results on the iPhone 15 Pro NPU, and (8) further discussion of the modality-routing mixture-of-experts. In addition, we also analyze the limitations of our work and outline directions for future research.

%


\section{More Details of the Proposed Method}

\subsection{Detailed Model Configurations}

We scale the proposed DualComp and DualComp-I models by varying the number of blocks and embedding dimensions, as these two factors primarily determine the model capacity and computational cost. The number of blocks is varied to control model depth, while the embedding dimension is increased to expand the feature space. Following \cite{rwkv7}, the hidden size of the MLP layers is set to $4\times$ the embedding dimension by default. Other model configurations are also aligned with \cite{rwkv7}. As shown in \cref{tab:structure}, we adopt six model configurations in total. Under the same configuration, DualComp has a larger model size than DualComp-I due to the inclusion of modality-specific parameters. 



\begin{table}[h]
    \centering
    \tabsize
    \belowrulesep=0pt
    \aboverulesep=0pt
    \renewcommand{\arraystretch}{1.0}
    \setlength{\tabcolsep}{8.5pt}
    \vspace{-6pt}
    \begin{tabular}{cc|p{2cm}c|c}
    \toprule
    \textbf{Blocks}
    & \textbf{Embed dim}
    & \textbf{Model}
    & \textbf{\#Params (M)}
    & \textbf{Dual-Modal}\\

    \midrule
    \multirow{2}{*}{2} & \multirow{2}{*}{96} 
    & DualComp & 0.3 & \checkmark\\
    & & DualComp-I & 0.2 & \ding{55}\\
    \midrule

    \multirow{2}{*}{2} & \multirow{2}{*}{192} 
    & DualComp & 1.2 & \checkmark\\
    & & DualComp-I & 0.8 & \ding{55}\\
    \midrule

    \multirow{2}{*}{2} & \multirow{2}{*}{384} 
    & DualComp & 4.8 & \checkmark\\
    & & DualComp-I & 3.2 & \ding{55}\\
    \midrule

    \multirow{2}{*}{2} & \multirow{2}{*}{720} 
    & DualComp & 16 & \checkmark\\
    & & DualComp-I & 12 & \ding{55}\\
    \midrule

    \multirow{2}{*}{3} & \multirow{2}{*}{1184} 
    & DualComp & 66 & \checkmark\\
    & & DualComp-I & 48 & \ding{55}\\
    \midrule

    \multirow{2}{*}{4} & \multirow{2}{*}{1456} 
    & DualComp & 130 & \checkmark\\
    & & DualComp-I & 96 & \ding{55}\\

    \bottomrule
    \end{tabular}
    \caption{\titlesize Model configurations across different model sizes and types.}
    \label{tab:structure}
\end{table}

\subsection{Detailed introduction of DualComp-I}

To further validate the effectiveness of our approach, we design a simplified single-modality variant, DualComp-I, for lossless image compression. This model removes the dual-modality proposals (modality-switching contextual learning and modality-routing mixture-of-experts) from DualComp, while preserving image tokenization and the reparameterization training strategy. The image tokenization vocabulary is fixed to 256. DualComp-I shares the same model configurations as DualComp but has fewer parameters (see \cref{tab:structure}), since it omits the modality-specific components. In this work, we do not introduce a dedicated variant for lightweight text-only compression, as similar proposals have already been explored in L3TC \cite{l3tc}.

DualComp-I is trained on the ImageNet dataset using the FusedAdam optimizer \cite{fusedadam}. We apply a cosine annealing learning rate scheduler \cite{scheduler} and train the model for 16 epochs. The learning rate starts at $1\times10^{-1}$ and decays to $2\times10^{-5}$. The training is performed on two NVIDIA A100 GPUs.

\subsection{Detailed Reparameterization Training Strategy}

\begin{wrapfigure}{r}{0.38\textwidth}
\centering
\vspace{-10pt}
\includegraphics[width=\linewidth]{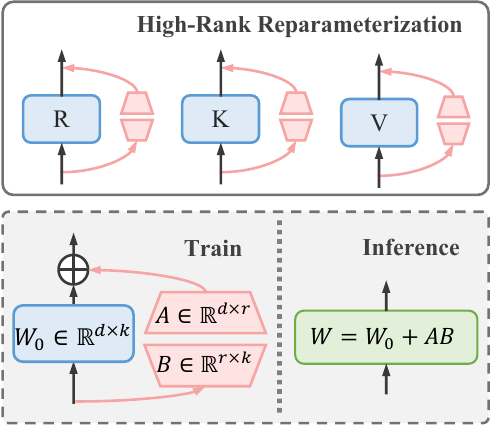}
\caption{\titlesize Illustration of the high-rank reparameterization training strategy.}
\label{fig:reparam}
\end{wrapfigure}

Following \cite{l3tc}, we adopt a high-rank reparameterization training strategy to enhance compression performance. During training, each R, K, and V layer in the Time Mixing module is augmented with an additional branch to improve learning capacity. At inference, these branches are merged back into the main path to keep the parameter count low. Further, we increase the expressiveness of these branches using matrix decomposition, as shown in \cref{fig:reparam}


Specifically, instead of using parallel branches like $1 \times 1$ convolutions or shortcuts as in \cite{reparam}, we follow the strategy in \cite{l3tc} and reparameterize each branch as a product of two high-rank matrices: $A \in \mathbb{R}^{d \times r}$ and $B \in \mathbb{R}^{r \times k}$, where the main-path's weight is $W_0 \in \mathbb{R}^{d \times k}$ and $r \gg d, k$. During training, the layer's output is the sum of the main branch and the bypass branch. These branches are merged at inference via structural reparameterization: $W = W_0 + A \times B$, resulting in a single-path structure that reduces both runtime and memory usage. Importantly, although the structure is simplified, the learned multi-branch parameters are preserved, maintaining high inference performance. 

In this paper, DualComp introduces modality-switching contextual learning to the Time Mixing module, hence the R, K, V projection layers are split into two sets corresponding to the two modalities. We apply the reparameterization training strategy to both sets, using a decomposition rank $r$ equal to $4\times$ embedding dimension following \cite{l3tc}.

\subsection{Detailed Modality-Routing Mixture-of-Experts}

As shown in Fig.2 of the main paper, DualComp replaces the large MLP in the final block with a lightweight MoE layer \cite{moe}. This layer is composed of a set of compact MLP experts and a trainable router $\text{G}$ that assigns input tokens to the appropriate experts.

For each input token $x_i$, the router computes a routing score $g_{i,e}$ for each expert $e$ via projections as:
\begin{equation}
    g_{i,e} = \frac{\exp({x_i\cdot \text{G}))}}{\sum_{e=1}^E \exp({x_i\cdot \text{G}))}}
\end{equation}
To promote diversity and avoid early expert collapse, optional Gaussian noise is added to the scores during training. Subsequently, experts with the top-$k$ highest scores are then activated to process token $x_i$. Each selected expert $e$ receives all tokens assigned to it and produces an output $e(x_i)$. The final output is obtained by aggregating the expert outputs using a weighted sum:
\begin{equation}
    \text{MoE}(x_i) = \sum_{e\in \text{top}-k}\hat{g}_{i,e}\cdot e(x_i)
\end{equation}
where $\hat{g}_{i,e}$ is the re-normalized routing score among the selected experts. Such a sparsely activated structure allows for expert specialization while keeping efficient parallel computation. 

To ensure balanced expert usage, we introduce an auxiliary load-balancing loss as:
\begin{equation}
\mathrm{CV}^2(\sum g_{i,e})+\mathrm{CV}^2(\sum \mathbb{I}(g_{i,e}>0))
\end{equation}
Here, $\mathrm{CV}^2$ denotes the squared coefficient of variation, calculated as the variance divided by the square of the mean. The first term encourages uniform expert importance, while the second promotes balanced token distribution across experts. This regularization helps avoid overloading a few experts and ensures more effective use of model capacity.



\begin{table*}[!tb]
    \belowrulesep=0pt
    \aboverulesep=0pt
    \centering
    \tabsize
    \renewcommand{\arraystretch}{0.98}
    \setlength{\tabcolsep}{2.35pt}
    \newcommand{\centerdash}[1]{\ifx#1-\multicolumn{1}{c}{-}\else#1\fi}
    \vspace{-6pt}
    \begin{tabular}{p{0.28cm}p{1.7cm}|p{1.22cm}cc|c|p{1.42cm}p{1.42cm}p{1.42cm}p{1.42cm}|c}
    \toprule
     & \multirow{2}{*}{\textbf{Compressor}} 
     & \multirow{2}{*}{\textbf{\#Params}$\downarrow$} 
     & \multirow{2}{*}{\textbf{MACs}$\downarrow$} 
     & \multirow{2}{*}{\makecell{\textbf{Speed}$\uparrow$\\ \textbf{(KB/s)}}}
     & \multirow{2}{*}{\textbf{\makecell{Dual\\Modal}}} 
     & \multicolumn{4}{c}{\textbf{bits/Byte$\downarrow$\quad [image]}}
     & \textbf{[text]} \\[1pt]
     \cmidrule{7-11}

     & & & & & 
     & \textbf{Kodak} & \textbf{CLIC-P} & \textbf{CLIC-M} & \textbf{DIV2K} & \textbf{enwik9} \\
    \midrule
    \multirow{9}{*}{\rotatebox{90}{\textbf{Classical}}} 
    & gzip$^\dag$ \cite{gzip}   & - & - & - & \checkmark 
    & 4.349 & 4.039 & 3.947 & 4.224 & 2.590\\
    & bzip2$^\dag$ \cite{bzip2} & - & - & - & \checkmark
    & 4.359 & 4.033 & 3.931 & 4.208 & 2.082 \\
    & zstd$^\dag$ \cite{zstd}   & - & - & - & \checkmark
    & 4.350 & 4.042 & 3.952 & 4.229 & 2.364 \\
    & FLIF$^\dag$ \cite{flif}   & - & - & - & \ding{55} 
    & 2.903 & 2.792 & 2.497 & 2.914 &  - \\
    & BPG$^\dag$ \cite{bpg}     & - & - & - & \ding{55} 
    & 3.382 & 3.144 & 2.855 & 3.285 &  - \\
    & WebP$^\dag$ \cite{webp}   & - & - & - & \ding{55} 
    & 3.205 & 3.006 & 2.774 & 3.176 &  - \\
    & JPEG-XL$^\dag$ \cite{jpegxl}  & - & - & - & \ding{55} 
    & 2.902 & 2.722 & 2.395 & 2.826 &  - \\
    & JPEG2000$^\dag$ \cite{jpeg2000} & - & - & - & \ding{55} 
    & 3.301 & 2.927 & 2.667 & 3.127 &  - \\
    & PNG$^\dag$ \cite{png}     & - & - & - & \ding{55} 
    & 4.349 (+54\%) & 4.041 (+72\%) & 3.952 (+90\%) & 4.229  (+68\%) &  - \\
    \cmidrule{1-11}

    \multirow{19}{*}{\rotatebox{90}{\textbf{Learning-based}}}
    & NNCP \cite{nncpv2:2021} & - & 187M & 1.6 & \ding{55} 
    & - & - & - & - & 0.853 \\
    & CMIX \cite{cmix:2023} & - & - & 4 & \ding{55} 
    & - & - & - & - & 0.879 \\
    & tszip \cite{tszip}  & 169M & 131M & 180$^\ddag$ & \ding{55} 
    & - & - & - & - &  1.083 \\
    & L3TC \cite{l3tc}  & 12M & 13.0M & 580$^\ddag$ & \ding{55} 
    & - & - & - & - &  1.280 \\
    
    & L3C \cite{l3c}     & 5M & 6.86M & 178 & \ding{55} 
    & 3.260  & 2.940 & 2.640 & 3.090 & - \\
    & RC \cite{rc}       & - & - & - & \ding{55} 
    & - & 2.930 & 2.540 & 3.080 & - \\
    & iVPF \cite{ivpf}   & - & - & - & \ding{55} 
    & - & 2.540 & 2.390 & 2.680 & - \\
    & iFlow \cite{iflow} & - & - & - & \ding{55} 
    & - & \underline{2.440} (+4\%) & 2.260 & 2.570 & - \\
    & DLPR \cite{dlpr}   & - & - & 640  & \ding{55} 
    & 2.860  & \underline{2.380} (+1\%) & \underline{2.160} (+4\%) & \underline{2.550} (+2\%) & - \\
    & P2LLM \cite{p2llm} & 8B ($1\times$) & - & 3$^\ddag$ & \ding{55} 
    & \underline{2.830} (0\%) 
    & \textbf{2.350} (0\%) 
    & \textbf{\color{darkblue} 2.080} (0\%) 
    & \textbf{\color{darkblue} 2.510} (0\%) 
    & - \\

    & $\text{Chinchilla}^{*}$ \cite{lmic} & 1B & - & - & \checkmark & - & - & - & - & 0.904 \\
    & $\text{Chinchilla}^{*}$ \cite{lmic} & 7B & - & - & \checkmark & - & - & - & - & \underline{0.816} \\
    & $\text{Chinchilla}^{*}$ \cite{lmic} & 70B & - & - & \checkmark & - & - & - & - & \textbf{\color{darkblue} 0.642} \\
    
    & $\text{Llama3}^{*\dag}$ \cite{lmic} & 1B & 927M & 9$^\ddag$ & \checkmark & 5.183 & 4.626 & 4.654 & 5.126 & 0.887 \\
    & $\text{Llama3}^{*\dag}$ \cite{lmic} & 3B & 2.39G & 5$^\ddag$ & \checkmark & 4.903 & 4.364 & 4.385 & 4.693 & \underline{0.798} \\
    & $\text{Llama3}^{*\dag}$ \cite{lmic} & 8B & 7.80G & 3$^\ddag$ & \checkmark & 4.862 & 4.290 & 4.234 & 4.378 & \textbf{0.722} \\

    & $\text{RWKV}^{*\dag}$ \cite{lmic}  & 169M & 159M  & 18$^\ddag$ & \checkmark & 5.334 & 4.830 & 4.879 & 5.190 & 1.158 \\
    & $\text{RWKV}^{*\dag}$ \cite{lmic}  & 1.5B & 1.41G  & 13$^\ddag$ & \checkmark & 4.954
    & 4.360 & 4.450 & 4.758 & 0.871 \\
    & $\text{RWKV}^{*\dag}$ \cite{lmic}  & 7B & 7.19G  & 5$^\ddag$ & \checkmark & 4.937 & 4.310 & 4.298 & 4.718 & \underline{0.774} \\
    \cmidrule{1-11}

    \multirow{11}{*}{\rotatebox{90}{\textbf{Proposed}}}
    & \cellcolor{lightblue}DualComp-I  
    & \cellcolor{lightblue}3.2M 
    & \cellcolor{lightblue}4.69M 
    & \cellcolor{lightblue}910$^\ddag$
    & \cellcolor{lightblue}\ding{55}
    & \cellcolor{lightblue}3.134
    & \cellcolor{lightblue}2.870 
    & \cellcolor{lightblue}2.628  
    & \cellcolor{lightblue}3.013 
    & \cellcolor{lightblue}- \\

    & \cellcolor{lightblue}DualComp-I  
    & \cellcolor{lightblue}12M 
    & \cellcolor{lightblue}11.6M 
    & \cellcolor{lightblue}614$^\ddag$
    & \cellcolor{lightblue}\ding{55}
    & \cellcolor{lightblue}\underline{2.823} (-0\%) 
    & \cellcolor{lightblue}2.674 
    & \cellcolor{lightblue}2.448  
    & \cellcolor{lightblue}2.862 
    & \cellcolor{lightblue}- \\
    
    & \cellcolor{lightblue}DualComp-I  
    & \cellcolor{lightblue}48M ({\color{darkblue}$\frac{1}{167}$}) 
    & \cellcolor{lightblue}33.3M 
    & \cellcolor{lightblue}360$^\ddag$
    & \cellcolor{lightblue}\ding{55} 
    & \cellcolor{lightblue}\textbf{2.693} ({\color{darkblue}-5\%}) 
    & \cellcolor{lightblue}2.455 
    & \cellcolor{lightblue}\underline{2.193} (+5\%) 
    & \cellcolor{lightblue}\underline{2.627} (+5\%)
    & \cellcolor{lightblue}- \\
    
    & \cellcolor{lightblue}DualComp-I   
    & \cellcolor{lightblue}96M ({\color{darkblue}$\frac{1}{83}$}) 
    & \cellcolor{lightblue}59.9M
    & \cellcolor{lightblue}317$^\ddag$
    & \cellcolor{lightblue}\ding{55}
    & \cellcolor{lightblue}\textbf{\color{darkblue} 2.571} ({\color{darkblue}-9\%}) 
    & \cellcolor{lightblue}\textbf{\color{darkblue} 2.350} (-0\%) 
    & \cellcolor{lightblue}\textbf{2.110} (+1\%) 
    & \cellcolor{lightblue}\textbf{2.547} (+2\%) 
    & \cellcolor{lightblue}- \\
    \cmidrule{2-11}

    & \cellcolor{lightpink}DualComp   
    & \cellcolor{lightpink}0.3M 
    & \cellcolor{lightpink}0.87M 
    & \cellcolor{lightpink}5264$^\ddag$
    & \cellcolor{lightpink}\color{darkblue}{\checkmark}
    & \cellcolor{lightpink}3.604 
    & \cellcolor{lightpink}3.330 
    & \cellcolor{lightpink}3.227
    & \cellcolor{lightpink}3.544 
    & \cellcolor{lightpink}1.360 \\

    & \cellcolor{lightpink}DualComp   
    & \cellcolor{lightpink}1.2M 
    & \cellcolor{lightpink}1.99M 
    & \cellcolor{lightpink}3560$^\ddag$
    & \cellcolor{lightpink}\color{darkblue}{\checkmark}
    & \cellcolor{lightpink}3.331 
    & \cellcolor{lightpink}3.042 
    & \cellcolor{lightpink}2.874
    & \cellcolor{lightpink}3.249 
    & \cellcolor{lightpink}1.332 \\

    & \cellcolor{lightpink}DualComp   
    & \cellcolor{lightpink}4.8M 
    & \cellcolor{lightpink}4.97M 
    & \cellcolor{lightpink}783$^\ddag$
    & \cellcolor{lightpink}\color{darkblue}{\checkmark}
    & \cellcolor{lightpink}3.138 
    & \cellcolor{lightpink}2.977 
    & \cellcolor{lightpink}2.809
    & \cellcolor{lightpink}3.083 
    & \cellcolor{lightpink}1.302 \\

    & \cellcolor{lightpink}DualComp   
    & \cellcolor{lightpink}16M 
    & \cellcolor{lightpink}12.6M 
    & \cellcolor{lightpink}550$^\ddag$
    & \cellcolor{lightpink}\color{darkblue}{\checkmark}
    & \cellcolor{lightpink}3.026 
    & \cellcolor{lightpink}2.691 
    & \cellcolor{lightpink}2.498
    & \cellcolor{lightpink}2.901 
    & \cellcolor{lightpink}1.218 \\
    
    & \cellcolor{lightpink}DualComp  
    & \cellcolor{lightpink}66M ({\color{darkblue}$\frac{1}{121}$}) 
    & \cellcolor{lightpink}36.0M 
    & \cellcolor{lightpink}280$^\ddag$
    & \cellcolor{lightpink}\color{darkblue}{\checkmark} 
    & \cellcolor{lightpink}2.889 
    & \cellcolor{lightpink}2.533 
    & \cellcolor{lightpink}2.370
    & \cellcolor{lightpink}2.741 
    & \cellcolor{lightpink}1.135 \\
    
    & \cellcolor{lightpink}DualComp  
    & \cellcolor{lightpink}130M ({\color{darkblue}$\frac{1}{60}$}) 
    & \cellcolor{lightpink}64.0M 
    & \cellcolor{lightpink}195$^\ddag$
    & \cellcolor{lightpink}\color{darkblue}{\checkmark}
    & \cellcolor{lightpink}\underline{2.834} (+0\%) 
    & \cellcolor{lightpink}\underline{2.364} (+0\%)  
    & \cellcolor{lightpink}\underline{2.155} (+3\%)
    & \cellcolor{lightpink}\underline{2.554} (+2\%)
    & \cellcolor{lightpink}1.107 \\
    
    \bottomrule
    \end{tabular}
    \caption{\titlesize Comprehensive lossless compression results on image and text datasets: best (bits/Byte) in \textbf{\color{darkblue}bold blue}, second in \textbf{bold}, and third to fifth \underline{underlined}. Image compression gains and model size ratios over the previous best method \cite{p2llm} are reported as percents (\%). $*$ denote pretrained LLMs. $\dag$ marks the methods we reprodeuced, while $\ddag$ presents the inference speeds measured on GPU@A100. All other unmarked values are claimed by their original papers.
}
    \label{tab:compare-more}
\end{table*}

\section{Extended Experiments}

\subsection{Details of the Dual-Modality Datasets}

Our datasets include both image and text modalities. The training set consists of 5,500 images from ImageNet \cite{imagenet} and 100MB of text from enwik8 \cite{enwik8}. Although the image and text training sets differ in size, we apply data augmentation to align their sample counts, ensuring each batch contains an equal number of image and text samples. Experiments show that the balanced 1:1 ratio leads to stable training, especially for those modality-shared parameters.


During evaluation, we use four common datasets to assess the image compression performance. (1) DIV2K \cite{div2k}: A high-quality super-resolution dataset containing one hundred 2K resolution images with diverse textures, commonly used for evaluating detail preservation in compression tasks. (2) CLIC-M \cite{clic}: 61 smartphone-captured high-resolution images (mostly 2K) featuring real-world scenes with noise and compression artifacts, reflecting mobile imaging challenges. (3) CLIC-P \cite{clic}: 41 images (mostly 2K) from DSLR/mirrorless cameras, providing professional-grade content with complex color gradients. (4) Kodak: The classic 24-image benchmark (768×512) containing film-scanned photos with balanced natural and synthetic content. 

For text evaluation, we use enwik9 \cite{enwik9}: The first 1GB of English Wikipedia XML dump, containing structured text with diverse vocabulary and mixed content types (articles, tables, markup).

\subsection{More Results on Lossless Compression Performance}

The main paper omits the detailed analysis of some results from \cite{lmic}. As shown in \cref{tab:compare-more}, the Chinchilla models \cite{chinchilla} are closed-source, so we only report their enwik9 \cite{enwik9} compression results from \cite{lmic}, with no results available for other datasets. It achieves the best text compression performance (0.642 bits/Byte) with 70B parameters. Herein, we also include the performance of smaller pretrained LLMs ( Llama3-1B \cite{llama3}, Llama3-3B, and RWKV-169M \cite{rwkv4}, RWKV-1.5B) using the method in \cite{lmic}, while our main paper reports only the pretrained Llama3-8B and RWKV-7B for a general comparison. These smaller models show poor compression performance on image datasets (e.g., pretrained Llama3-1B leads to 5.183 and 5.126 bits/Byte on Kodak and DIV2K, respectively) and moderate performance on text (e.g., pretrained RWKV-169M yields 1.158 bits/Byte on enwik9).


Besides, the results of our small models (both DualComp and DualComp-I) are also provided in \cref{tab:compare-more}. Though these small models offer moderate compression performance on image and text data, they can reach up to 5.3 MB/s inference speed on an NVIDIA A100 GPU with a batch size of 128, making them well-suited for low-latency scenarios where the compression ratio is less critical.

\begin{figure}[t]
    \centering
    \includegraphics[width=0.325\linewidth]{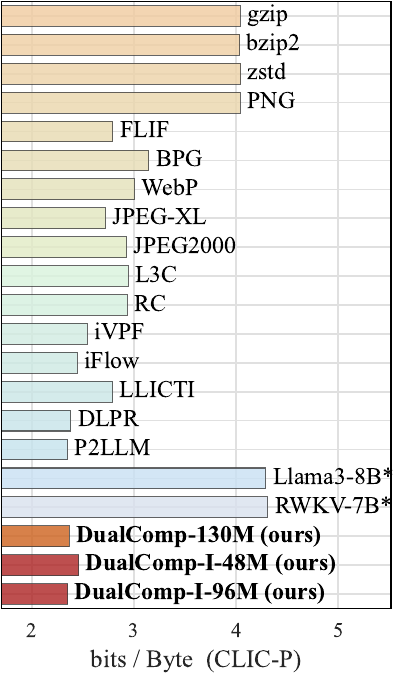}
    \hfill
    \includegraphics[width=0.325\linewidth]{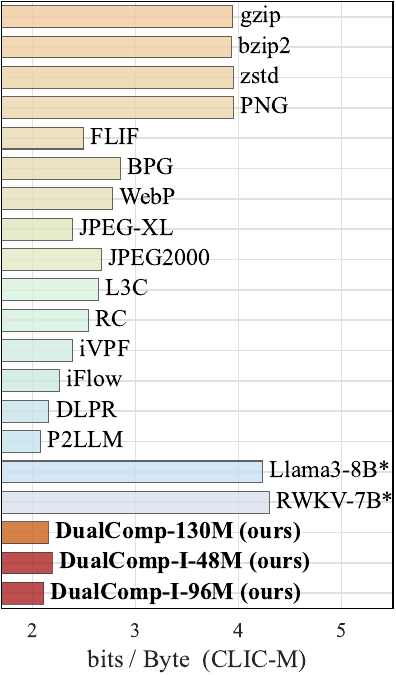}
    \hfill
    \includegraphics[width=0.325\linewidth]{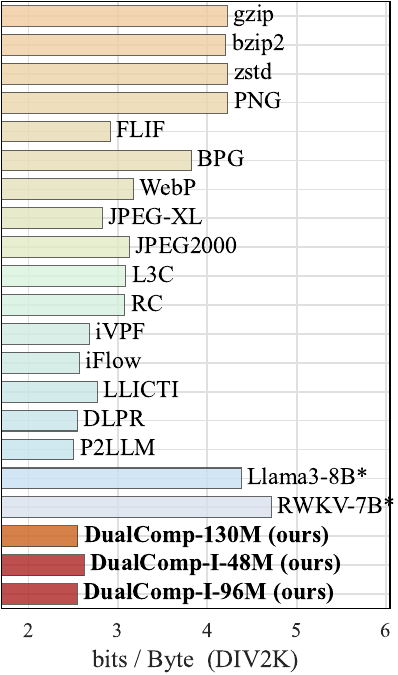}
    \caption{\titlesize Lossless compression results (bits/Byte) on CLIC-P, CLIC-M, and DIV2K datasets. DualComp-I achieves SOTA performance on CLIC-P and ranks second on CLIC-M and DIV2K. DualComp ranks third on these datasets.}
    \label{fig:teaser-more}
\end{figure}

Further, we also present comparisons of our method's compression performance on the other three image datasets (CLIC-P, CLIC-M, DIV2K), as shown in \cref{fig:teaser-more}. These figures correspond to the first plot in the main paper and offer a clear visual comparison in terms of bits/Byte. It can be seen that our DualComp-I achieves SOTA results (2.350 bits/Byte) on the CLIC-P dataset with only 1.2\% parameters of the previous best method \cite{p2llm}. It also ranks second on both CLIC-M and DIV2K. Besides, despite being trained on dual-modality data and having two-thirds of its parameters shared across modalities, our DualComp model still ranks third or fourth across all datasets. Our methods perform better on the Kodak dataset, likely because we divide images into small $16\times16\times3$ patches, and the training set consists mostly of low-resolution images.

\subsection{More Results on Compression Complexity}

\begin{figure}
    \centering
    \includegraphics[width=0.48\linewidth]{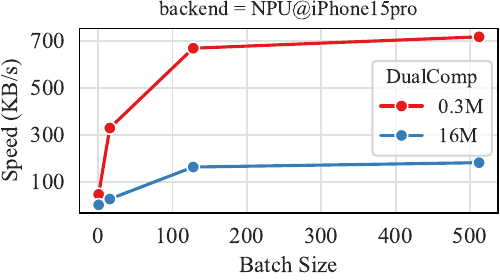}
    \hfill
    \includegraphics[width=0.48\linewidth]{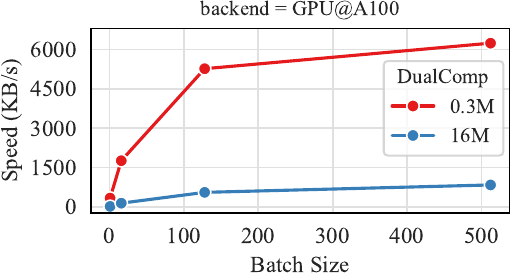}
    \caption{\titlesize DualComp's inference speed (KB/s) on an iPhone 15 Pro's NPU (left) and a NVIDIA A100 GPU (right) at varying batch sizes (1, 16, 128, and 512). Two model sizes (0.3M and 16M) are used for illustration.}
    \label{fig:iphone-speeds}
\end{figure}

In the main paper (Section 4.3), we report the inference speed of our methods on a MacBook Air CPU and an NVIDIA A100 GPU. Herein, we further present results on the iPhone 15 Pro’s NPU (Apple Neural Engine), as illustrated in \cref{tab:speed-more} and \cref{fig:iphone-speeds}. We convert models to CoreML \cite{coreml} packages and use Xcode software to measure the inference speed.

\cref{tab:speed-more} compares the inference speed of our models with existing methods. We evaluate three representative model sizes for DualComp-I (0.2M, 12M, 96M) and DualComp (0.3M, 16M, 130M), representing small, medium, and large configurations. Small and medium models are tested on all three hardware platforms, while the larger 96M and 130M models are evaluated only on the A100 GPU. All models use a batch size of 128. It can be seen that DualComp-0.3M achieves 134 KB/s on the MacBook CPU and 688 KB/s on the iPhone NPU, enabling real-time inference on mobile devices. The larger models (12M and 16M) also reach near real-time inference (about 200KB/s) on iPhone 15 Pro. The DualComp-I models are even smaller and achieve faster inference speed exceeding 1.1 MB/s on NPU@iPhone. When running on GPU servers, our methods achieve inference speeds comparable to lightweight compressors such as \cite{l3c, l3tc, dlpr}, and are several hundred to several thousand times faster than those LLM-based approaches \cite{lmic,p2llm}. 


\begin{table}[t]
    \belowrulesep=0pt
    \aboverulesep=0pt
    \centering
    \tabsize
    \renewcommand{\arraystretch}{0.98}
    \setlength{\tabcolsep}{7.5pt}
    \newcommand{\centerdash}[1]{\ifx#1-\multicolumn{1}{c}{-}\else#1\fi}
    \vspace{-6pt}
    \begin{tabular}{p{0.35cm}p{2cm}|p{1.2cm}p{1cm}|p{1.8cm}p{1.8cm}p{2cm}}
    \toprule
    \multicolumn{2}{c|}{\multirow{2}{*}{\textbf{Compressor}}} 
    & \multirow{2}{*}{\textbf{\#Params} $\downarrow$} 
    & \multirow{2}{*}{\textbf{MACs} $\downarrow$} 
    & \multicolumn{3}{c}{\textbf{Speed $\uparrow$ (KB/s)}} \\
    \cmidrule{5-7} 
    \multicolumn{2}{c|}{} & & & \textbf{CPU@MacBook} & \textbf{NPU@iPhone} & \textbf{GPU@server} \\
   \midrule
   \multirow{14}{*}{\rotatebox{90}{\textbf{Learning-based}}}
   & NNCP \cite{nncpv2:2021} & - & 187M & - & - & 1.6\\
   & CMIX \cite{cmix:2023}   & - & - & - & - & 4\\
   & tszip \cite{tszip}      & 169M & 131M & - & - & 180\\
   & L3TC \cite{l3tc}        & 12M  & 13.0M & - & - & 580$^\ddag$\\
   & L3C \cite{l3c}          & 5M   & 6.86M & - & - & 178\\
   & DLPR \cite{dlpr}        & -    & - & - & - & 640\\
   & P2LLM \cite{p2llm}      & 8B ($1\times$) & - & - & - & 3$^\ddag$ ($1\times$)\\
   & Llama3$^{*\dag}$ \cite{lmic} & 1B & 927M & - & - & 9$^\ddag$\\
   & Llama3$^{*\dag}$ \cite{lmic} & 3B & 2.39G & - & - & 5$^\ddag$\\
   & Llama3$^{*\dag}$ \cite{lmic} & 8B & 7.80G & - & - & 3$^\ddag$\\
   & RWKV$^{*\dag}$ \cite{lmic} & 169M & 159M & - & - & 18$^\ddag$\\
   & RWKV$^{*\dag}$ \cite{lmic} & 1.5B & 1.41G & - & - & 13$^\ddag$\\
   & RWKV$^{*\dag}$ \cite{lmic} & 7B   & 7.19G & - & - & 5$^\ddag$\\
   \midrule
   
   \multirow{6.5}{*}{\rotatebox{90}{\textbf{Proposed}}}
   & \cellcolor{lightblue}DualComp-I 
   & \cellcolor{lightblue}0.2M
   & \cellcolor{lightblue}0.85M
   & \cellcolor{lightblue}151
   & \cellcolor{lightblue}1163
   & \cellcolor{lightblue}5808$^\ddag$ ($1936\times$)\\
   
   & \cellcolor{lightblue}DualComp-I 
   & \cellcolor{lightblue}12M  
   & \cellcolor{lightblue}11.57M
   & \cellcolor{lightblue}36  
   & \cellcolor{lightblue}227  
   & \cellcolor{lightblue}614$^\ddag$\\
   
   & \cellcolor{lightblue}DualComp-I 
   & \cellcolor{lightblue}96M ($\frac{1}{83}$)
   & \cellcolor{lightblue}59.9M
   & \cellcolor{lightblue}-   
   & \cellcolor{lightblue}-    
   & \cellcolor{lightblue}317$^\ddag$  ($106\times$)\\
   \cmidrule{2-7}
   
   & \cellcolor{lightpink}DualComp   
   & \cellcolor{lightpink}0.3M 
   & \cellcolor{lightpink}0.87M 
   & \cellcolor{lightpink}134 
   & \cellcolor{lightpink}668
   & \cellcolor{lightpink}5264$^\ddag$ ($1755\times$)\\
   
   & \cellcolor{lightpink}DualComp   
   & \cellcolor{lightpink}16M  
   & \cellcolor{lightpink}12.55M 
   & \cellcolor{lightpink}22  
   & \cellcolor{lightpink}163 
   & \cellcolor{lightpink}550$^\ddag$\\
   
   & \cellcolor{lightpink}DualComp   
   & \cellcolor{lightpink}130M ($\frac{1}{60}$)
   & \cellcolor{lightpink}64.0M 
   & \cellcolor{lightpink}-   
   & \cellcolor{lightpink}-   
   & \cellcolor{lightpink}195$^\ddag$ ($65\times$)\\
    \bottomrule
    \end{tabular}
    \caption{\titlesize Inference speed (KB/s) comparison of learning-based lossless compressors. The proposed DualComp-I and DualComp models are evaluated on three platforms (MacBook Air CPU, iPhone 15 Pro NPU, and NVIDIA A100 GPU) with a batch size of 128. The large models (96M and 130M) models merely run on A100 GPU. Baseline methods are tested only on GPU servers, with  $\ddag$ indicating the results on A100. $*$ marks the pretrained LLMs, $\dag$ denotes methods reproduced by us, and the unmarked results are taken from their original papers. 
}
    \label{tab:speed-more}
\end{table}

Besides, the main paper reports inference speed scaling with batch size only on a MacBook Air CPU. In \cref{fig:iphone-speeds}, we extend this analysis to the iPhone 15 Pro’s NPU and the A100 GPU, evaluating batch sizes of 1, 16, 128, and 512. It can be seen that across all devices, inference speed increases with batch size and begins to saturate beyond 128. At batch size 512, the DualComp-0.3M and DualComp-16M models reach about 6.3 MB/s and 0.83 MB/s, respectively, on the A100 GPU.

\subsection{More Results on Modality-Routing Mixture-of-Experts}

\begin{figure}[t]
    \centering
    \includegraphics[width=\linewidth]{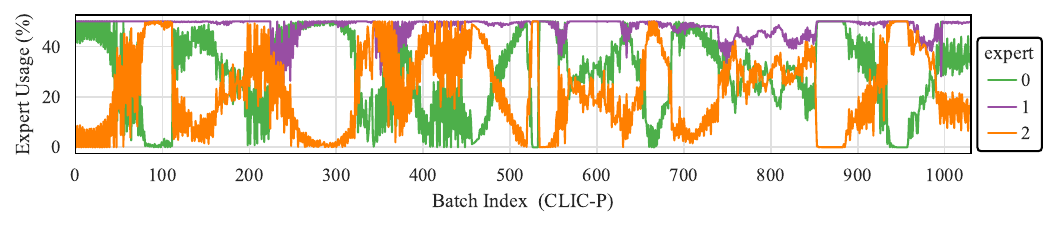}\\
    \includegraphics[width=\linewidth]{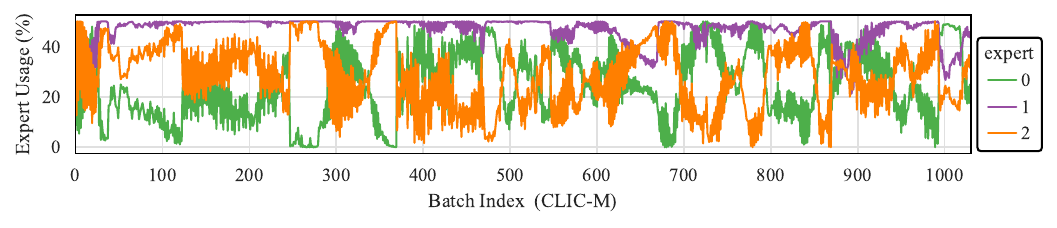}\\
    \includegraphics[width=\linewidth]{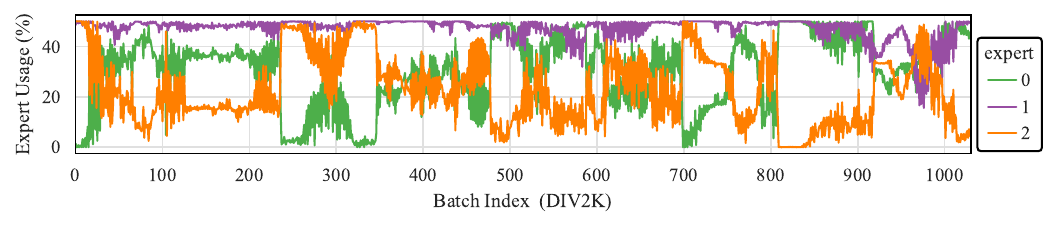}
    \caption{\titlesize Expert usage per batch when compressing CLIC-P, CLIC-M, and DIV2K using DualComp-16M. Expert 1 is most frequently used expert across image patches, likely because it captures more general image features.}
    \label{fig:experts-more}
\end{figure}

Due to space limitations, the main paper uses the Kodak dataset as an example to present MoE expert usage when compressing images. Herein, we extend the analyses and show per batch expert utilization when compressing CLIC-P, CLIC-M, and DIV2K datasets, as depicted in \cref{fig:experts-more}. The x-axis corresponds to the batch index (with batch size equaling 128), while the y-axis presents the percent of expert usage (\%) within each batch. 

It can be observed that different images show distinct expert usage patterns, which are noticeably more complex than those in text compression (see Fig.4 in the main paper). This complexity likely comes from the longer temporal dependencies created by flattening 2D image patches into 1D sequences. However, among all experts, the expert-1 is used most frequently across all image patches, indicating that it may have learned to extract more general image features. Usage of the other two experts is more random, with the expert-2 being used more often for text tokens (see Fig.4 in the main paper).

\section{Future Works}

Currently, our DualComp framework supports only dual-modality compression. In future work, we plan to extend it to more challenging modalities such as audio and video, with the goal of building a more unified lossless compressor. Notably, our model structure is highly extensible. The most straightforward approach to supporting new modalities is to add a corresponding tokenizer, along with branches for modality-switching contextual learning and modality-routing MoE adaptation. Additional techniques will also be investigated to enhance the model’s multi-modality understanding and enable more lightweight deployment on edge devices.

Moreover, the current DualComp model allocates equal parameters to both modalities for simplicity. Future work will investigate more efficient allocation strategies that better account for the compression difficulty of each modality. We will also explore more effective training methods to address the challenge of balancing performance across modalities.

\end{document}